\pdfoutput=1

\documentclass[11pt]{article}

\usepackage{ACL2023}

\usepackage{times}
\usepackage{latexsym}

\usepackage[T1]{fontenc}

\usepackage[utf8]{inputenc}

\usepackage{microtype}

\usepackage{inconsolata}
\usepackage{graphicx}
\usepackage{subcaption}

\usepackage{awesomebox}
\usepackage{xcolor}
\usepackage{float}
\usepackage{listings}
\usepackage{booktabs}
\usepackage{todonotes}

\definecolor{separation}{HTML}{E6F4Fb}
\definecolor{all}{HTML}{E1F2DB}
\definecolor{brown}{HTML}{FDF1E4}
\definecolor{brown2}{HTML}{E9E4F6}
\definecolor{dark}{HTML}{C0E3FA}

\definecolor{snowviolet}{HTML}{7447C7}
\definecolor{snoworange}{HTML}{F2A24E}
\definecolor{snownavy}{HTML}{28567C}

\DeclareRobustCommand{\shl}[3]{
  \begingroup\setlength{\fboxsep}{2pt}%
  \colorbox{#1}{{\hspace*{2pt}\vphantom{Ay}#2\hspace*{2pt}}}%
  \endgroup
}

\title{In Case You Missed It:\\ARC `Challenge' Is Not That Challenging}

\author{Łukasz Borchmann \\
  Snowflake AI Research \\
  \texttt{lukasz.borchmann@snowflake.com}}

\begin{document}
\maketitle

\begin{abstract}
ARC Challenge appears more difficult than ARC Easy for modern LLMs primarily due to an evaluation setup that prevents direct comparison of answer choices rather than inherent complexity. Although some researchers have quietly shifted to a more appropriate scheme over the last year, the implications of this change have yet to be widely acknowledged. We highlight this overlooked shift, show how similar evaluation practices falsely imply reasoning deficits in other benchmarks, and demonstrate that fairer methods dramatically reduce performance gaps (e.g. on SIQA) and even yield superhuman results (OpenBookQA). In doing so, we reveal how evaluation shapes perceived difficulty and offer guidelines to ensure that multiple-choice evaluations accurately reflect actual model capabilities.
\end{abstract}

\section{Introduction}

A substantial set of benchmarks regularly employed in LLM testing consists of multiple-choice problems, commonly considered in a setup where each provided option is scored under the model, and the one with the highest likelihood is compared against the gold standard to determine accuracy. This refers, among others, to popular evaluators of MMLU \cite{hendrycks2021measuringmassivemultitasklanguage}, ARC Easy and Challenge \cite{clark2018thinksolvedquestionanswering}, BoolQ \cite{clark2019boolqexploringsurprisingdifficulty}, RACE \cite{lai2017racelargescalereadingcomprehension}, OpenBookQA \cite{mihaylov2018suitarmorconductelectricity}, PIQA \cite{DBLP:journals/corr/abs-1911-11641}, SIQA \cite{sap2019socialiqacommonsensereasoningsocial}, COPA \cite{Gordon2011ChoiceOP}, and HellaSwag \cite{zellers2019hellaswagmachinereallyfinish}.

\begin{figure}[ht]
    \centering
    \includegraphics[width=\linewidth]{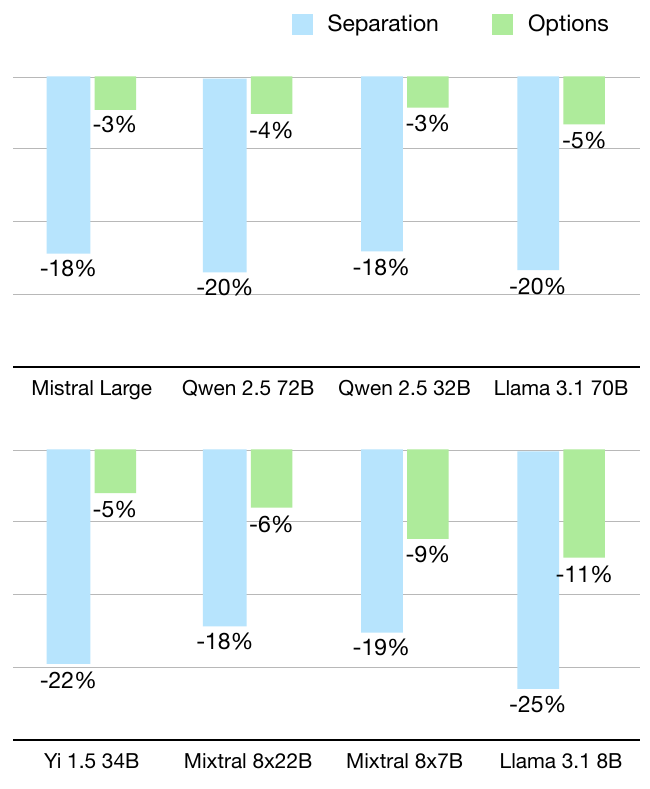}
    \caption{Difference between {\color{snownavy}\textbf{ARC Challenge}} and {\color{snownavy}\textbf{ARC Easy}} accuracies when considering each answer separately compared to seeing all options. The gap is vastly reduced, up to six times in this comparison.}
    \label{fig:difference}
\end{figure}

\begin{figure*}
    \centering
    \includegraphics[width=0.85\linewidth]{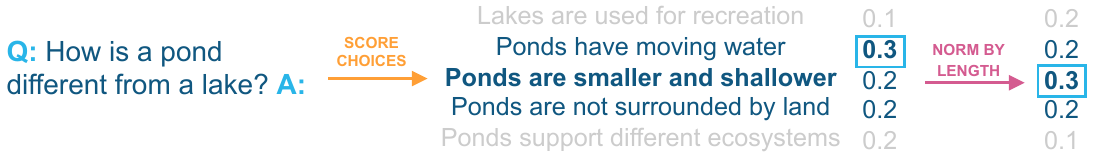}
    \caption{Model considers particular choices in \shl{separation}{separation}{} without knowing the alternative (prompt includes only the question). Because options may vary in length, it is a good practice to normalize them \cite{norm}.}\label{fig:arc}
\end{figure*}

\begin{figure*}
    \centering
    \includegraphics[width=0.6\linewidth]{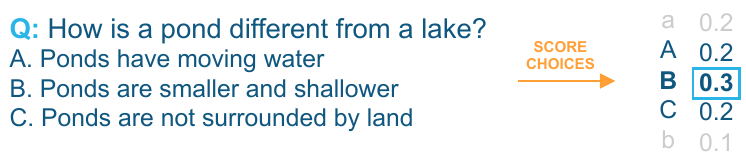}
    \caption{Model sees the context of all possible \shl{all}{options}{} in the prompt. Because all of the options are single letters (likely single tokens), scores require no normalization.}\label{fig:mmlu}
\end{figure*}

Details of this setup differ but generally follow one of the two conventions. 
Under one convention, the model considers each candidate answer in \shl{separation}{separation}{}, without alternative options displayed (Figure~\ref{fig:arc}), while under the other, the model sees all candidate \shl{all}{options}{} together in the prompt (Figure~\ref{fig:mmlu}).
We argue that the first setup is commonly overused and rarely preferred since it does not simulate the natural reasoning context in which multiple choices are compared directly. Importantly, it introduces a false notion of how challenging a particular problem is, as switching from the first to the second might result in a 35\% improvement in model accuracy, as shown in Section~\ref{sec:impact} experiments.

\subsection{Hardly answerable in separation}\label{sec:hardly}

Consider the question, `Which of these items contains only a solution?' Given the option `a jar of pickles,' confronting a single item with a question and assessing whether pickles fulfill the definition of the solution suffices. They do not, so this option is incorrect. The question can be addressed under both evaluation setups because it does not require the availability of other options, such as `a can of mixed fruit.'

In contrast, some questions inherently demand comparative evaluation: let us think about `Which of these most likely has the \textit{greatest} mass?' and the option `puppy.' 
This question’s answer cannot be determined without comparing the mass of the 'puppy' to the masses of all other provided options. It is the greatest compared to `chicken' or `lizard' but not in the context of `horse' or `elephant.' Though it can work to some extent, relying on the likelihood assigned in \shl{separation}{separation}{} to each of the animals is an unreasonable way of determining the heaviest one. It feels natural to provide the model with the \shl{all}{options}{} to choose from instead because it allows the model to directly compare and contextualize choices, reflecting a more authentic reasoning process. This aspect, however, is commonly overlooked.

Specifically, such `hardly answerable in separation' questions are prevalent in ARC datasets, constituting 21\% of ARC Easy and 31\% of ARC Challenge (see Appendix~\ref{appendix:hardly}). 
Despite this fact, it is widespread to evaluate them without seeing all of the options simultaneously \cite[\textit{inter alia}]{touvron2023llamaopenefficientfoundation,touvron2023llama2openfoundation,jiang2023mistral7b,peng2023rwkvreinventingrnnstransformer,ai2024yiopenfoundationmodels,gemmateam2024gemmaopenmodelsbased}.

\section{Impact on evaluation results}\label{sec:impact}

\begin{figure*}
    \centering
    \includegraphics[width=0.95\linewidth]{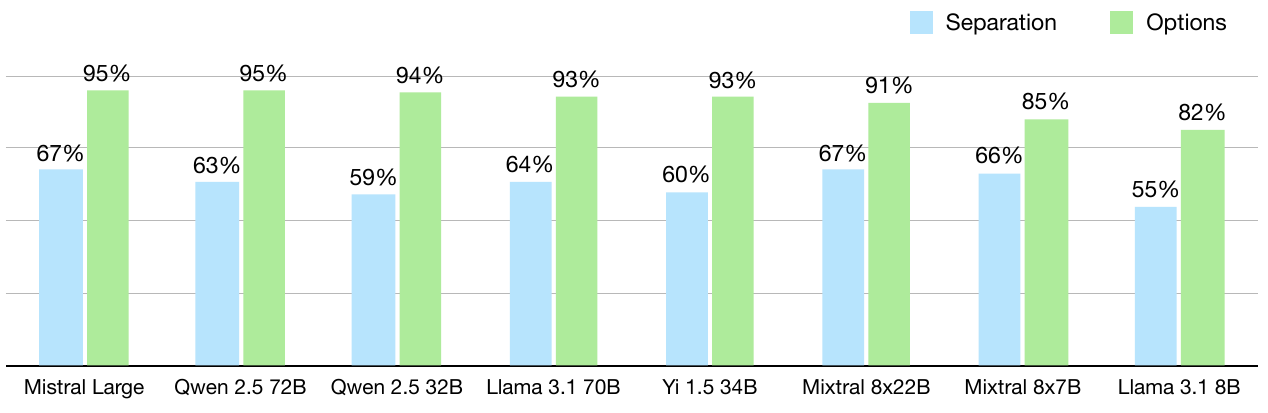}
    \caption{{\color{snownavy}\textbf{ARC Challenge}} evaluation results depending on whether the model sees other options or considers each answer separately. Differences reach up to 35\%, and assumed setup impacts model rankings.}
    \label{fig:challenge}
\end{figure*}

Figure~\ref{fig:challenge} shows the difference in model accuracy when options are presented in isolation versus all at once.
Not surprisingly, different setups hugely change the evaluation results, partly because of the vast presence of `hardly answerable in separation' questions and partially because such a setup, equivalent to what human test takers see, doesn't introduce unnecessary obstacles.

For example, switching from \shl{separation}{separation}{} to \shl{all}{options}{} improves the Llama 3.1 70B ARC Challenge accuracy from 64\% to 93\%, rendering this ARC subset significantly less challenging.
Moreover, since the procedure change has a much higher impact on ARC Challenge than on ARC Easy, switching reduces the accuracy gap between these subsets as much as six-fold (Figure~\ref{fig:difference}). 
These findings suggest that the previously perceived difficulty was primarily an artifact of the evaluation method rather than the tasks' complexity.

The difference seems somewhat known in the LLM community, but not broadly, and needs to be stated explicitly. E.g. concerning the Llama family, authors seem to silently switch from \shl{separation}{separation}{} to \shl{all}{options}{} between Llama~2 and Llama~3, similar to Mistral between Mixtral 8x7B and Mixtral 8x22B, or DeepSeek before their V2 (detailed assessment available in Appendix~\ref{appendix:claims}).

\section{Are other benchmarks affected?}

\begin{figure}[t]
    \centering
    \includegraphics[width=\linewidth]{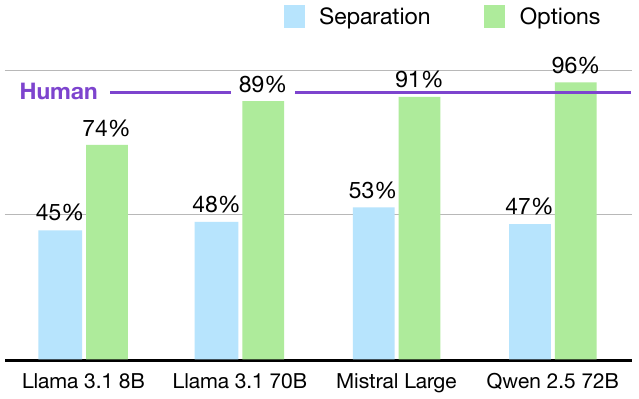}
    \caption{{\color{snowviolet}\textbf{OpenBookQA}} evaluation results depending on whether the model sees other options or considers each answer separately. In a setup with options, current models outperform human test takers.}
    \label{fig:openbookqa}
\end{figure}

Yes. Analogous changes in evaluation procedures would vastly improve OpenBookQA scores. Concerning Llama 3.1 70B, one can achieve improvement from 48\% to 89\% (see Figure~\ref{fig:openbookqa}). For some reason, most authors who switched from \shl{separation}{separation}{} to \shl{all}{options}{} in ARC evaluation did not follow on some other multi-choice problems. 

\begin{figure}[t]
    \centering
    \includegraphics[width=\linewidth]{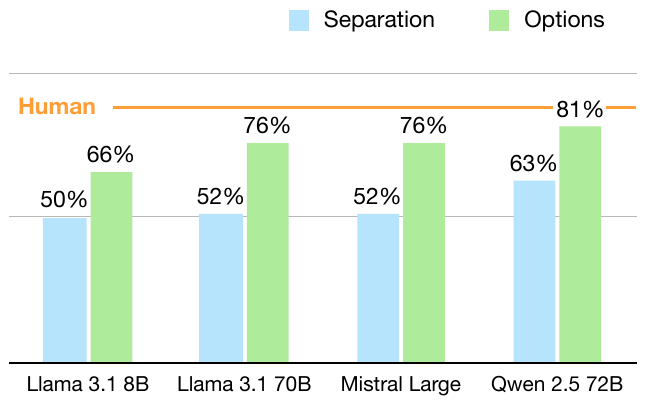}
    \caption{{\color{snoworange}\textbf{SIQA}} evaluation results depending on whether the model sees other options or considers each answer separately. Reformulation leads to up to 24\% improvement.}
    \label{fig:siqa}
\end{figure}

If they switched, they could notice that OpenBookQA is essentially solved, as current models achieve scores above human performance (92\% compared to 96\% of Qwen 2.5 72B).

In the case of SIQA, reformulation leads to a 24\% increase in Llama 70B accuracy. However, the best models perform 5\% below the human baseline, suggesting room for improvement (Figure~\ref{fig:siqa}).

These dramatic increases, however, call into question previous interpretations of model capability on both SIQA and OpenBookQA.

\section{Why does it matter?}

We argue that the benchmark's challenge should result from the inherent complexity of the knowledge or reasoning required, not its formulation or evaluation procedure.

The \shl{separation}{separation}{} setup is unnecessarily complicated and not consistent with how humans would approach the multi-choice problem, leading to existing assessments of human performance being incompatible. For example, the fact that strong LLMs perform 30\% worse than humans on SIQA doesn't mean they are deficient in commonsense reasoning about social situations if under \shl{all}{options}{} the difference largely disappears. This mismatch can falsely suggest deficits in reasoning capabilities that are not truly present.

Notably, the gap between LLMs and humans on SIQA has been previously used to argue that LLMs might lack social intelligence and social commonsense \cite{sap-etal-2022-neural}.

\section{Suggestions for multi-choice eval}

There are many arguments for using the \shl{all}{options}{} for the evaluation of multi-choice QA problems. We have already described a few, including the presence of `hardly answerable in separation' questions and the fact it is consistent with the usual approach to assessing human performance, as humans naturally consider all choices in a single context. 

Other benefits include enabling compatible evaluation in a likelihood and generative manner, allowing one to obtain comparable scores with LLMs behind closed and limited APIs. Moreover, it eliminates the need to decide which normalization method to use when aggregating scores from several option tokens, which is, to some extent, arbitrary and impacts model ranking. 

Nevertheless, it is not preferred in all cases commonly considered under the likelihood-scoring evaluation scheme.

\subsection{Why likelihood scoring in the first place?}

Likelihood-based scoring is a natural choice for problems from pure language modeling, Winograd schemas, or fill-in-the-gap setups such as encountered in LAMBADA \cite{paperno2016lambadadatasetwordprediction}, HellaSwag \cite{DBLP:journals/corr/abs-1911-11641}, or WindoGrande \cite{sakaguchi2019winograndeadversarialwinogradschema} datasets.

For other problems, it is effectively a variant of constrained decoding, that is, the model is restricted to selecting from given candidate options rather than generating open-ended text. It guarantees that models will not emit CoTs before answering the question and removes the need for output postprocessing, such as extracting the letter associated with the selected option and normalizing its casing. Moreover, it allows us to obtain meaningful results with base models, e.g. intermediate checkpoints from LLMs' self-supervised pretraining since we are constraining the output to one of the most probable options under the model.

\subsection{To show, or not to show options}

Suppose the options are of equal length, and it is not helpful to consider them simultaneously. This is the case when we deal with a straightforward yes/no response, and no comparative reasoning is necessary, as in the BoolQ dataset \cite{clark2019boolqexploringsurprisingdifficulty}. In similar scenarios, there are no arguments for dropping \shl{separation}{separation}{} in favor of \shl{all}{options}{}.

We are in the position that the \shl{all}{options}{} variant is preferred if there is a risk of a `hardly answerable in separation' question presence (Section~\ref{sec:hardly}) or it simply makes it easier to consider all of the options at once because it ensures the model can leverage direct comparisons. This seems to be the case for virtually all other multi-choice QA problems, such as MMLU \cite{hendrycks2021measuringmassivemultitasklanguage}, ARC \cite{clark2018thinksolvedquestionanswering}, OpenBookQA \cite{mihaylov2018suitarmorconductelectricity}, PIQA \cite{DBLP:journals/corr/abs-1911-11641}, or SIQA \cite{sap2019socialiqacommonsensereasoningsocial}.
In fact, most similar problems are already being evaluated in the \shl{all}{options}{} scheme.

Arguably, \shl{all}{options}{} has some possible or actual disadvantages: the order of the presented choices might impact the evaluation results (e.g. models might bias toward the first listed choice), it might be easier to exploit pattern recognition, and the setup requires slightly more compute. Nevertheless, we consider these to be outweighed by benefits and recommend broad use of the \shl{all}{options}{} scheme.


\section*{Limitations}
Despite our recommendations and the benefits they entail, several limitations and uncertainties exist in identifying precisely how evaluation methods were employed in previously reported results.

Firstly, because authors of LLMs' technical reports rarely or never report such details, our assessment of which of the \shl{all}{options}{} and \shl{separation}{separation}{} they employed is based on attempting to replicate reported accuracy scores under both setups and observing which condition aligns best (Appendix~\ref{appendix:claims}). Fortunately, given the magnitude of the observed differences, the performance gap is so large that one can differentiate between the mentioned approaches with a high degree of certainty.

Secondly, in a search for the \shl{separation}{separation}{} overuse candidates, we mainly relied on intuition. We considered the most popular benchmarks due to their widespread use and availability of performance data, later confirming the intuition experimentally. Though it was a tale of ARC, OpenBookQA, and SIQA, many other widely used benchmarks may benefit from revisiting their evaluation setup.

Finally, it could be a blog post, but it feels better to write a PDF.

\section{Summary}

We draw the community's attention to shifting from evaluating answers in isolation to evaluating them alongside all other options. Over the last year, such a change happened in the reported ARC Challenge and ARC Easy scores, vastly impacting their evaluation results. After discussing the implications, we considered whether other popular benchmarks might undergo similar reformulation, identifying OpenBookQA and SIQA as candidates. In the former, recent models outperform humans, even though there is a room of 40\% between humans and LLMs in the widespread setup. The fact that the gap drastically narrows under the all-options evaluation method highlights how the testing format can distort perceived difficulty.

We concluded with a guideline for evaluating multi-choice problems, arguing that the setup where the model sees all options is preferred over considering each answer separately, except for casual or masked language modeling problems. 

\clearpage

\bibliography{anthology,custom}
\bibliographystyle{acl_natbib}

\clearpage
\appendix
\begin{table*}[ht]
    \centering
    \small
    \def\arraystretch{1.25}
    \begin{tabular}{lcccl}
        \toprule
        Model & Reported & Measured \shl{separation}{s}{}  /\shl{all}{o}{} & Assessment \\ \midrule 
         Llama 65B \cite{touvron2023llamaopenefficientfoundation} & 56.0 & 55.6 / 70.2 & \shl{separation}{separation}{} \\
         Llama 2 70B \cite{touvron2023llama2openfoundation} & 57.4 & 57.4 / 79.6 & \shl{separation}{separation}{} \\
         Llama 3 70B \cite{grattafiori2024llama3herdmodels} & 92.9 & 64.2 / 91.3 & \shl{all}{options}{} \\
         Mistral 7B \cite{jiang2023mistral7b} & 55.5 & 54.1 / 74.6 & \shl{separation}{separation}{} \\
         Mixtral 8x7B \cite{jiang2024mixtralexperts} & 59.7 & 59.9 / 83.3 & \shl{separation}{separation}{} \\
         Mixtral 8x22B \cite{mixtral22} & ~~91.3$^\dagger$ & 70.7 / 91.8 &  \shl{all}{options}{}\\
         DeepSeek 67B \cite{deepseekai2024deepseekllmscalingopensource} & 59.0 & 60.1 / 84.6 & \shl{all}{options}{}\\
         DeepSeek V2 \cite{deepseekai2024deepseekv2strongeconomicalefficient} & ~~92.4$^\dagger$ & 70.3 / 92.2 & \shl{all}{options}{} \\
         Qwen 14B \cite{bai2023qwentechnicalreport} & 84.4 & 47.3 / 86.6 & \shl{all}{options}{}\\
         Yi 6B \cite{ai2024yiopenfoundationmodels} & ~~50.3$^\dagger$ & 55.7 / 80.5 & \shl{separation}{separation}{} \\
         Gemma 7B \cite{gemmateam2024gemmaopenmodelsbased} & 53.2 & 53.2 / 79.0 & \shl{separation}{separation}{} \\
         Gemma 2 27B \cite{gemmateam2024gemma2improvingopen} & 71.4 & 65.8 / 90.0 & \shl{separation}{separation}{} \\
        \bottomrule
    \end{tabular}
    \caption{Measured and reported {\color{snownavy}\textbf{ARC Challenge}} scores with our assessment of the setup used by authors. The 25-shot prompting used in contrast to the 0-shot is denoted by $\dagger$ (in the case authors use such a setup in their report).}
    \label{tab:claims}
\end{table*}
\begin{table*}[ht]
    \centering
    \small
    \def\arraystretch{1.25}
    \begin{tabular}{lcccl}
        \toprule
        Model & Reported & Measured \shl{separation}{s}{}  /\shl{all}{o}{} & Assessment \\ \midrule 
         Llama 65B \cite{touvron2023llamaopenefficientfoundation} & 52.3 & 50.3 / 60.1 & \shl{separation}{separation}{} \\
         Llama 2 70B \cite{touvron2023llama2openfoundation} & 50.7 & 50.8 / 66.9 & \shl{separation}{separation}{} \\
         Llama 3 70B \cite{grattafiori2024llama3herdmodels} & 52.2 & 51.2 / 72.9 &  \shl{separation}{separation}{} \\
         Mistral 7B \cite{jiang2023mistral7b} & ~~---$^{\diamond}$ & 50.9 / 62.4 & --- \\
         Mixtral 8x7B \cite{jiang2024mixtralexperts} & ~~---$^{\diamond}$ & 49.4 / 65.1 & --- \\
         Mixtral 8x22B \cite{mixtral22} & --- & 51.1 / 67.3 & --- \\
         DeepSeek 67B \cite{deepseekai2024deepseekllmscalingopensource} & --- & 51.6 / 61.6 & --- \\
         DeepSeek V2 \cite{deepseekai2024deepseekv2strongeconomicalefficient} & --- & 52.2 / 70.0 & --- \\
         Qwen 14B \cite{bai2023qwentechnicalreport} & 77.9 & 56.2 / 78.6 & \shl{all}{options}{} \\
         Yi 6B \cite{ai2024yiopenfoundationmodels} & --- & 52.5 / 71.0 & --- \\
         Gemma 7B \cite{gemmateam2024gemmaopenmodelsbased} & 51.8 & 51.8 / 60.0 & \shl{separation}{separation}{} \\
         Gemma 2 27B \cite{gemmateam2024gemma2improvingopen} & 53.7 & 58.3 / 70.0  & \shl{separation}{separation}{} \\
        \bottomrule
    \end{tabular}
    \caption{Measured and reported {\color{snoworange}\textbf{SIQA}} scores with our assessment of the setup used by authors. Some authors do not directly report scores but average them with other commonsense reasoning problems (denoted by $^{\diamond}$), making our assessment unlikely to succeed.}
    \label{tab:siqa-claims}
\end{table*}
\begin{table*}[ht]
    \centering
    \small
    \def\arraystretch{1.25}
    \begin{tabular}{lcccl}
        \toprule
        Model & Reported & Measured \shl{separation}{s}{} / \shl{all}{o}{} /\shl{brown}{s$_\mathrm{b}$}{} / \shl{brown2}{o$_\mathrm{b}$}{} & Assessment \\ \midrule 
         Llama 65B \cite{touvron2023llamaopenefficientfoundation} & 60.2 & 47.0 / 59.0 / 60.2 / 56.2 & \shl{brown}{separation $_\mathrm{b}$}{} \\
         Llama 2 70B \cite{touvron2023llama2openfoundation} & 60.2 & 48.8 / 73.0 / 60.0 / 65.8 & \shl{brown}{separation $_\mathrm{b}$}{} \\
         Llama 3 70B \cite{grattafiori2024llama3herdmodels} & 47.6 & 48.6 / 88.4 / 59.4 / 88.5 & \shl{separation}{separation}{} \\
         Mistral 7B \cite{jiang2023mistral7b} & ~~---$^{\diamond}$ &  44.2 / 71.6 / 55.0 / 57.8 & --- \\
         Mixtral 8x7B \cite{jiang2024mixtralexperts} & ~~---$^{\diamond}$ & 47.0 / 80.2 / 55.2 / 78.0 & --- \\
         Mixtral 8x22B \cite{mixtral22} & --- & 49.6 / 81.6 / 61.2 / 78.4 & --- \\
         DeepSeek 67B \cite{deepseekai2024deepseekllmscalingopensource} & 60.2 & 47.6 / 76.6 / 62.0 / 76.2 & \shl{brown}{separation $_\mathrm{b}$}{} \\
         DeepSeek V2 \cite{deepseekai2024deepseekv2strongeconomicalefficient} & --- & 38.6 / 82.8 / 62.4 / 84.2 & --- \\
         Qwen 14B \cite{bai2023qwentechnicalreport} & --- & 43.8 / 87.0 / 54.6 / 79.8 & --- \\
         Yi 6B \cite{ai2024yiopenfoundationmodels} & ~~---$^{\diamond}$ & 40.4 / 68.2 / 53.6 / 67.6 & --- \\
         Gemma 7B \cite{gemmateam2024gemmaopenmodelsbased} & --- & 44.8 / 65.2 / 58.2 / 65.8 & --- \\
         Gemma 2 27B \cite{gemmateam2024gemma2improvingopen} & --- & 47.6 / 83.0 / 59.8 / 81.4  & --- \\
        \bottomrule
    \end{tabular}
    \caption{Measured and reported {\color{snowviolet}\textbf{OpenBookQA}} scores with our assessment of the setup used by authors. Some authors do not directly report scores but average them with other commonsense reasoning problems (denoted by $^{\diamond}$). 
    }
    \label{tab:openbookqa-claims}
\end{table*}

\section{Claiming setup used by other authors}\label{appendix:claims}
As authors of LLM technical reports rarely or never provide such details, we redo their evaluations in \shl{all}{options}{} and \shl{separation}{separation}{} setups. If the reported score is in the same ballpark as one of these, and visibly distant from the other one, we claim they used the first.
This assessment is backed by the notion that no other change in the prompt could cause a 20\%+ improvement in ARC Challenge scores. The exception could be using a generative setup with CoT for some heavy reasoners, but we do not suspect authors to use CoT if they are not reporting this because it would be a serious flaw.

The results of this analysis for the ARC Challenge are presented in Table~\ref{tab:claims}. Prompt `reverse engineering' becomes more troublesome in the context of SIQA and OpenBookQA datasets (Table~\ref{tab:siqa-claims}-\ref{tab:openbookqa-claims}) as some authors do not directly report scores but average them with other commonsense reasoning problems. We're not claiming any setup for these.

Finally, some authors tackling OpenBookQA followed \citet{DBLP:journals/corr/abs-2005-14165} in normalizing the likelihood by the likelihood of the completion given `Answer:' as context. To address this possibility, we introduce two additional variants referred to as \shl{brown}{separation $_\mathrm{b}$}{} and \shl{brown2}{options $_\mathrm{b}$}{}.

\section{Estimating number of questions hardly answerable in separation}
\label{appendix:hardly}

To determine whether questions are answerable given a single option or require the context of other options, we process them in batches of 20 using \texttt{gpt-4o-2024-11-20} model and the following prompt with few-shot examples:
\vspace{1em}
\begin{lstlisting}
Consider the question, "Which of these items contains only a solution?" Given the option "a jar of pickles," confronting a single item with a question and assessing whether pickles fulfill the definition of the solution suffices. They do not, so this option is incorrect. 

Now let us think about "Which of these most likely has the greatest mass?" and the option "puppy." It can be considered only with other options because it is the greatest compared to "chicken" or "lizard" but not in the context of "horse" or "elephant".

These questions represent two classes of questions: "answerable without other options" and "unanswerable without other options".

Other examples of "answerable without other options" are:
- Kerry made a simple flashlight. She recorded the following statements in her lab book. Which statement is an inference? (Answerable, because it suffices to compare options against the definition of inference)
- A scientist on a field trip discovered a new organism. She examined its cells under a microscope and observed several different structures, including a nucleus, a cell wall, and some chloroplasts. This organism would correctly be classified in which of the following kingdoms? (Answerable, because it can be answered by deciding if the kingdom provided in the option can be associated with having a nucleus, a cell wall, and chloroplasts)
- Many types of motion occur in our solar system. Which type of motion describes one Earth year? (Answerable, because it suffices to validate if the motion describes one year or not)
- When trees develop leaves in the spring, 10 changes occur on the forest floor. Why does the development of leaves cause changes on the forest floor? (Answerable, because it is enough to verify if a particular option described the possible cause of change)
- Using a softball bat to hit a softball is an example of using which simple machine? (Answerable, because all one needs to do is to check if the described simple machine is the explanation of how a softball bat works)
- Which is a statement about climate? (Answerable, because it is possible to verify a single option against the climate definition)
- How do word processors on computers benefit most students? (Answerable, because it can be answered in separation whether most students benefit from this feature of the word processor)
- Photosynthesis occurs in which of these organisms? (Answerable because it suffices to check if the organism mentioned in the option performs photosynthesis)
- Which two theories of Moon formation propose that much or all of the material comprising the Moon came from Earth? (Because it suffices to validate if both theories mentioned in a single option describe the Moon as formed from Earth material)
- Plants and animals are composed of organic compounds. Which of the following are the common elements found in organic compounds? (Answerable, because it suffices to check if the option consists of compounds appearing in both plants and animals)

Other examples of "unanswerable without other options" are:
- A ball is dropped from different heights. When the ball is dropped from the highest height, it makes the greatest noise or vibration when it lands on the ground. What is the best explanation for the ball making the greatest noise? (Unanswerable, because in order to choose the best explanation, one needs to consider several explanations)
- If an experiment results in data that do not support the hypothesis, what is the most likely step to take next? (Unanswerable, because in order to choose the most likely step, one needs to consider the less likely alternative)
- When an igneous intrusion comes into contact with surrounding rock, the surrounding rock will (Unanswerable, because one can easily verify if an option describes the possible outcome of contact with surrounding rock)
- A research scientist writes a paper on the initial regrowth of a forest after a fire has damaged the entire ecosystem. Which title would be best for the paper? (Unanswerable, because it is impossible to decide the best title without comparing it to other titles)
- Jessica wants to see cells in an oak tree leaf. Which tool is best for Jessica to use to see the cells? (Unanswerable, because choosing the best tool depends on the set of tools considered and is ambiguous without a complete list of options considered)
- Which factor is most likely to cause the number of rabbits living in an area to increase? (Unanswerable, because choosing the most likely case requires checking all of the causes under consideration)

Now classify the following statements either as "unanswerable" or "answerable" in separation.

Answer in a form of JSONL file containing "question", "category", and "explanation" keys.

Questions to classify:
[List of 20 questions and choices]
\end{lstlisting}

\noindent The model returned batches of JSONL, such as:
\begin{lstlisting}
{"question": "Which best describes the structure of an atom?", "category": "unanswerable", "explanation": "Determining the best description requires comparing all options to identify the most accurate one."}
{"question": "Which is a statement about climate?", "category": "answerable", "explanation": "It is possible to verify each option against the definition of climate to determine the correct answer."}
{"question": "During which activity should a student wear goggles?", "category": "answerable", "explanation": "It suffices to check if the activity described in the option requires goggles for safety."}
{"question": "Which natural event occurs with the most frequency?", "category": "unanswerable", "explanation": "Determining the most frequent event requires comparing the frequency of all listed events."}
\end{lstlisting}

\noindent During this procedure, we estimated the percentage as 21\% for ARC Easy and 31\% for ARC Challenge.

\section{Evaluation details}

All evaluations were conducted using \texttt{lm\_eval} \texttt{1170ef9} \cite{eval-harness}. We used HF implementations and base variants of models (exact versions in Table~\ref{tab:models}) with either default prompts and \texttt{acc\_norm} metric or prompts outlined below. 

Inferences were performed with bf16 precision, flash attention (whenever available), and dynamic batch size, using \texttt{transformers} 4.47.0 and \texttt{torch}  2.5.1 on eight NVIDIA H100 GPUs.

\begin{table}[ht]
    \centering
    \small
    \begin{tabular}{ll}
        \toprule
        Model \\
        \midrule
        \texttt{huggyllama/llama-65b} \\
        \texttt{meta-llama/Llama-2-70b-hf} \\
        \texttt{meta-llama/Meta-Llama-3-70B} \\
        \texttt{mistralai/Mistral-7B-v0.1} \\
        \texttt{mistralai/Mixtral-8x7B-v0.1} \\
        \texttt{mistralai/Mixtral-8x22B-v0.1} \\
        \texttt{deepseek-ai/deepseek-llm-67b-base} \\
        \texttt{deepseek-ai/DeepSeek-V2} \\
        \texttt{Qwen/Qwen-14B} \\
        \texttt{01-ai/Yi-6B} \\
        \texttt{google/gemma-7b} \\
        \texttt{google/gemma-2-27b} \\
        \bottomrule
    \end{tabular}
    \caption{Exact variants of models used for evaluation.}
    \label{tab:models}
\end{table}

Concerning ARC Easy and Challenge datasets, for the \shl{separation}{separation}{} setup, we follow the standard \texttt{lm\_eval} configuration with:

\begin{lstlisting}
doc_to_text: "Question: {{question}}\nAnswer:"
doc_to_target: "{{choices.label.index(answerKey)}}"
doc_to_choice: "{{choices.text}}"
\end{lstlisting}

\noindent In contrast, for the \shl{all}{options}{} setup, we use:
\begin{lstlisting}
doc_to_text: !function arc_utils.doc_to_text
doc_to_target: "{{choices.label.index(answerKey)}}"
doc_to_choice: "{{choices.label}}"
\end{lstlisting}
\noindent with \texttt{doc\_to\_text()} defined as:
\begin{lstlisting}
def doc_to_text(doc):
    prompt = "Question: " + doc["question"] + "\nOptions:\n"
    for l, t in zip(doc["choices"]["label"], doc["choices"]["text"]):
        prompt += l + '. ' + t + '\n'
    prompt += "Answer: "
    return prompt
\end{lstlisting}

\noindent Analogous changes were introduced to OpenBookQA and SIQA templates.

\end{document}